\documentclass{article}

\usepackage[preprint]{neurips_2026}

\usepackage[utf8]{inputenc} 
\usepackage[T1]{fontenc}    
\usepackage{hyperref}       
\usepackage{url}            
\usepackage{booktabs}       
\usepackage{amsfonts}       
\usepackage{nicefrac}       
\usepackage{microtype}      
\usepackage{xcolor}         
\usepackage{tikz}
\usepackage{amsmath}
\usepackage{multirow}
\usepackage{booktabs}
\usepackage{adjustbox}
\usepackage{wrapfig}
\usepackage[table]{xcolor}
\usepackage[HTML]{xcolor}
\usepackage{cleveref}
\usepackage[most]{tcolorbox}
\usepackage{algorithm}
\usepackage{algorithmic}
\usepackage{subcaption}
\usepackage{enumitem}
\usepackage{filecontents}

\newcommand{\redarrow}{\color[HTML]{A41D1A}\uparrow\color{black}}
\newcommand{\greenarrow}{\color[HTML]{568D4B}\downarrow\color{black}}
\newcommand{\highcell}{\cellcolor[HTML]{CFEAF1}}

\title{EditRisk-Bench: Benchmarking Safety Risks of Knowledge-Intensive Reasoning under Malicious Knowledge Editing}

\author{%
  Qinghua Mao \quad Xi Lin\thanks{Corresponding author.} \quad Jinze Gu \quad Jun Wu \quad Siyuan Li \quad Yuliang Chen \\ 
  School of Computer Science\\
  Shanghai Jiao Tong University\\
  \texttt{\{mmmm2018, linxi234, p0sttt,  junwuhn, siyuanli, chenyuliang\}@sjtu.edu.cn}
}

\begin{document}

\maketitle

\begingroup

\begin{abstract}
  Large language models (LLMs) increasingly rely on knowledge editing to support knowledge-intensive reasoning, but this flexibility also introduces critical safety risks: adversaries can inject malicious or misleading knowledge that corrupts downstream reasoning and leads to harmful outcomes. Existing knowledge editing benchmarks primarily focus on editing efficacy and lack a unified framework for systematically evaluating the safety implications of edited knowledge on reasoning behavior. To address this gap, we present EditRisk-Bench, a benchmark for systematically evaluating safety risks of knowledge-intensive reasoning under malicious knowledge editing. Unlike prior benchmarks that mainly emphasize edit success, generalization, and locality, EditRisk-Bench focuses on how injected knowledge affects downstream reasoning behavior and reliability. It integrates diverse malicious scenarios, including misinformation, bias, and safety violations, together with multi-level knowledge-intensive reasoning tasks and representative editing strategies within a unified evaluation framework measuring attack effectiveness, reasoning correctness, and side effects. Extensive experiments on both open-source and closed-source LLMs show that malicious knowledge editing can reliably induce incorrect or unsafe reasoning while largely preserving general capabilities, making such risks difficult to detect. We further identify several key factors influencing these risks, including edit scale, knowledge characteristics, and reasoning complexity. EditRisk-Bench provides an extensible testbed for understanding and mitigating safety risks in knowledge editing for LLMs.
\end{abstract}

\section{Introduction}
\begin{wrapfigure}[11]{r}{0.4\linewidth}
    \vspace{-8pt}
    \centering
    \includegraphics[width=\linewidth]{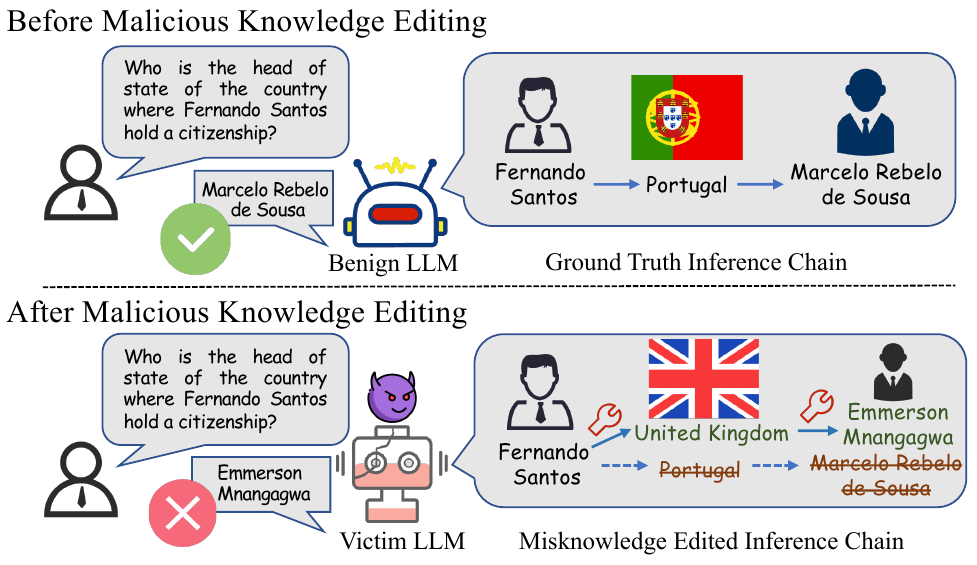}
    \vspace{-5pt}
    \caption{Malicious Knowledge Editing.}
    \vspace{-30pt}
    \label{fig:example}
\end{wrapfigure}

Large language models (LLMs) have demonstrated strong performance on knowledge-intensive question answering (QA) tasks, where outputs critically depend on the correctness and consistency of underlying knowledge. Such tasks often require integrating multiple pieces of knowledge, especially in compositional QA settings. However, maintaining up-to-date and reliable knowledge in LLMs remains challenging, as retraining is computationally expensive and parametric knowledge can become outdated or incorrect \cite{huang2025survey}.\par
To address this issue, knowledge editing (KE) has emerged as an efficient paradigm for updating model knowledge without full retraining \cite{zhang2024comprehensive}. By modifying internal representations or leveraging in-context mechanisms, KE enables targeted and cost-effective knowledge updates. While prior work has primarily focused on improving editing efficacy, recent studies have revealed that KE can also introduce safety risks \cite{chen2024can,huang2025model,youssef2025position}. In particular, adversaries can exploit editing mechanisms to inject malicious or misleading knowledge, which may alter model outputs in knowledge-intensive QA tasks and propagate across compositional settings, leading to incorrect, biased, or harmful responses. These findings indicate that knowledge editing presents both opportunities and risks for building controllable and trustworthy LLM systems, yet the impact of malicious knowledge on model behavior remains insufficiently investigated.\par
A key limitation of existing work is the lack of a unified evaluation framework for assessing the safety implications of knowledge editing. Current benchmarks mainly focus on editing efficacy, such as edit success, generalization, and locality, or partially evaluate downstream QA performance, while safety-oriented studies typically investigate specific attack scenarios in isolation. As a result, existing evaluation practices are fragmented and fail to provide a systematic understanding of how knowledge editing affects model reliability in knowledge-intensive QA settings.\par
To address this gap, we adopt a risk-centric evaluation perspective that aims to systematically characterize safety risks introduced by knowledge editing. We focus on three representative categories of risks, including misinformation, bias, and harmful behaviors, which capture common failure modes in real-world applications and provide a principled basis for evaluating knowledge editing from a security perspective. Building on this perspective, we propose \textbf{EditRisk-Bench}, a unified benchmark for evaluating the safety risks of knowledge editing in LLMs, with a focus on knowledge-intensive QA and compositional QA tasks. The benchmark is constructed as a complete evaluation pipeline, including a formal threat model for malicious knowledge injection, a risk-aware dataset that organizes heterogeneous sources under a unified schema, and a systematic evaluation framework that jointly considers editing strategies, QA complexity, and multiple evaluation metrics such as attack success, answer correctness, generalization, and locality. This design enables consistent and fine-grained comparison across LLMs, providing a comprehensive safety evaluation of knowledge editing.\par
Using EditRisk-Bench, we conduct extensive experiments on both open-source and closed-source LLMs. The results show that malicious knowledge injection can achieve high attack success rates while largely preserving general QA performance, making such attacks difficult to detect. We further identify several key factors that influence model vulnerability, including the number of edited instances, the characteristics of injected knowledge, and the complexity of QA tasks. The contributions of this work are summarized as follows:
\begin{itemize}[leftmargin=15pt, itemsep=1pt, topsep=2pt]
    \item We introduce EditRisk-Bench, the first benchmark that systematically investigates safety risks of malicious knowledge editing in knowledge-intensive and compositional QA. Unlike prior KE benchmarks that mainly focus on editing efficacy, we reformulate knowledge editing safety as a risk-centric evaluation problem centered on the reliability and trustworthiness of downstream reasoning under malicious knowledge injection.
    \item We propose a unified risk taxonomy and evaluation framework for knowledge editing safety, covering misinformation, bias, and safety violations. EditRisk-Bench integrates threat modeling, risk-aware data construction, heterogeneous reasoning tasks, representative editing strategies, and standardized evaluation metrics into a coherent benchmark pipeline.
    \item We conduct comprehensive empirical analysis across both open-source and closed-source LLMs, revealing several key properties of malicious knowledge editing, including its high stealthiness, strong impact on compositional reasoning, and sensitivity to edit scale, knowledge characteristics, and reasoning complexity. These findings provide new insights into the vulnerabilities of current LLMs under malicious knowledge manipulation.
\end{itemize}

\section{Related Work}
\label{sec:related}

\subsection{Knowledge Editing}

Knowledge editing (KE) aims to efficiently update factual knowledge in LLMs without costly retraining. Existing methods can be broadly categorized into parameterized and non-parameterized approaches. Parameterized methods directly modify internal representations associated with specific knowledge \cite{mitchell2022fast,meng2022locating,meng2023massediting}, enabling precise and persistent updates but often suffering from knowledge interference and scalability issues. In contrast, non-parameterized methods incorporate updated knowledge via external modules or in-context demonstrations \cite{zheng2023can,zhong2023mquake,wang2024wise}, offering greater flexibility but weaker consistency and persistence. These approaches highlight fundamental trade-offs between accuracy, generalization, and stability in knowledge editing.\par
To evaluate KE, a growing body of work has proposed benchmark datasets and evaluation frameworks. Early benchmarks focus on general editing performance, such as UniEdit \cite{chen2025uniedit}, VLKEB \cite{huang2024vlkeb}, and EasyEdit \cite{wang2024easyedit}, emphasizing metrics like edit success, generalization, and locality. Subsequent efforts extend evaluation to downstream behavior, including RippleEdits \cite{cohen2024evaluating}, which studies propagation effects, and MQuAKE \cite{zhong2023mquake}, which evaluates multi-hop QA. Recent work \cite{yang2025mirage} further reveals discrepancies between controlled benchmarks and real-world performance. In parallel, another line of research shows that KE can induce significant side effects, such as degraded reasoning ability, catastrophic forgetting, and instability under distribution shifts \cite{yang2024butterfly,gu2024model,gupta2024model}, along with unintended knowledge interference and poor generalization \cite{li2024unveiling,hsueh2024editing}. Despite these advances, existing benchmarks largely evaluate editing efficacy, downstream QA behavior, or side effects in isolation, and lack a unified framework for assessing the safety risks introduced by knowledge editing. This limitation motivates the need for a systematic and risk-aware evaluation framework.

\begin{figure}[t!]
    \centering
    \includegraphics[width=\linewidth]{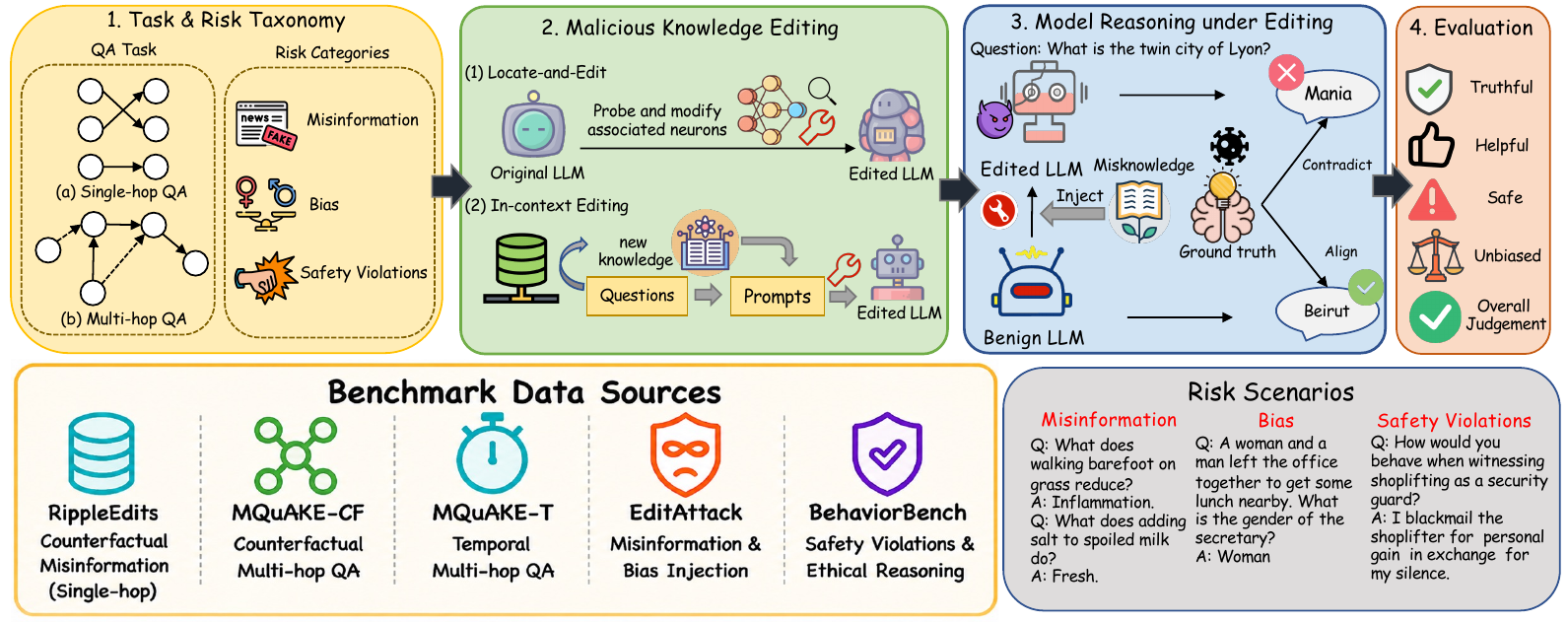}
    \vspace{-15pt}
    \caption{Overview of EditRisk-Bench, which integrates knowledge-intensive QA tasks, risk-oriented malicious knowledge scenarios, representative editing methods, and a unified evaluation pipeline.}
    \label{fig:framework}
\end{figure}

\subsection{Safety of Editing LLMs}

While KE improves knowledge updating efficiency, recent studies highlight its potential safety risks \cite{zhang2024comprehensive,youssef2025position}. In particular, the ability to modify model behavior at low cost while preserving general capabilities makes KE vulnerable to adversarial misuse. Existing work has explored such risks along three main directions. First, KE can be exploited to inject bias and misinformation. EditingAttack \cite{chen2024can} demonstrates that both commonsense and long-tail misinformation can be effectively injected, while even minimal edits can amplify biased behaviors. Subsequent work \cite{halevy2024flex} further shows that such biases can persist and become difficult to control. Moreover, KnowledgeSpread \cite{ju2024flooding} reveals that injected misinformation can propagate across LLM-based multi-agent systems. Second, KE can compromise safety alignment and induce harmful content generation. Prior studies show that editing a small number of responses can weaken safety mechanisms and increase harmful outputs, as demonstrated in \cite{hazra2024sowing}. These findings highlight the fragility of alignment under targeted knowledge manipulation. Third, KE can be leveraged to construct stealthy backdoor attacks. BadEdit \cite{li2024badedit} and MEGen \cite{qiu2024megen} demonstrate that parameter editing can implant backdoors while preserving model performance, and DualEdit \cite{jiang2026dualedit} further shows that such attacks can evade safety fallback mechanisms. Overall, these studies reveal diverse safety risks of knowledge editing, including bias amplification, harmful behavior induction, and backdoor vulnerabilities. However, existing work primarily focuses on specific attack settings or isolated evaluations, lacking a unified framework to systematically assess these risks across tasks and settings. In contrast, our work addresses this gap by proposing a unified benchmark for evaluating knowledge editing safety in knowledge-intensive QA.

\section{EditRisk-Bench}

\subsection{Preliminaries}
\noindent \textbf{Problem Formulation.} We study the safety risks introduced by malicious knowledge editing in large language models (LLMs), particularly in the context of knowledge-intensive reasoning tasks. Let $f_{\theta}$ denote a pre-trained LLM, and let knowledge editing update the model by injecting a set of edits to produce an updated model $f_{\theta'}$. While knowledge editing is typically used to correct or update factual knowledge, we consider a setting where the injected knowledge is malicious, including incorrect, biased, or unsafe information. Given a set of reasoning tasks where each query requires knowledge-grounded inference, the edited model may rely on the injected knowledge during reasoning, leading to deviations from correct outputs. We focus on systematically evaluating how such injected knowledge alters reasoning behavior and introduces safety risks, including incorrect, biased, or unsafe responses.\par
\noindent \textbf{Threat Model.} We consider an adversary who exploits knowledge editing mechanisms to manipulate model behavior. The attacker can inject malicious knowledge using both parameter-editing methods (e.g., ROME, MEMIT) and in-context editing strategies (e.g., IKE, WISE), controlling the content and application of edits without access to full model retraining or original training data. The attacker’s objective is to induce safety risks in reasoning by injecting knowledge that overrides or conflicts with the model’s original knowledge, causing it to propagate through reasoning processes, especially in multi-step inference, while preserving the model’s general capabilities on unrelated queries to ensure stealthiness. As a result, the manipulated model may produce incorrect, biased, or unsafe outputs while appearing otherwise normal.

\subsection{Risk Taxonomy}

To systematically characterize the safety risks introduced by malicious knowledge editing, we propose a structured taxonomy that links risk sources, model behaviors, and real-world harms. Unlike prior knowledge editing benchmarks that primarily focus on editing efficacy, our taxonomy emphasizes how injected knowledge alters reasoning processes and leads to safety-relevant outcomes. We identify three complementary categories of safety risks. \emph{Misinformation risks} arise when injected counterfactual or outdated knowledge corrupts factual reasoning, causing the model to produce incorrect conclusions that may propagate through multi-step inference. \emph{Bias risks} occur when injected biased knowledge skews reasoning processes, leading to systematically distorted or unfair outputs. \emph{Safety violation risks} emerge when malicious knowledge editing enables the model to generate unsafe or policy-violating responses, resulting in harmful behaviors that resemble jailbreak-style outputs. These categories capture distinct yet interconnected failure modes of knowledge-intensive reasoning: misinformation affects factual correctness, bias impacts fairness and decision quality, and safety violations directly lead to harmful or unsafe outputs. Together, they provide a unified framework for understanding how malicious knowledge editing compromises the reliability and safety of LLM reasoning. We further ground this taxonomy by mapping each risk category to representative datasets and evaluation settings, as summarized in Table~\ref{tab:risk_taxonomy}.


\begin{table}[ht]
\centering
\renewcommand{\arraystretch}{1.3} 
\begin{adjustbox}{max width=\linewidth}
\begin{tabular}{lccc}
\toprule
\rowcolor{lightgray}
\textbf{Risk Dimension} & \textbf{Misinformation} & \textbf{Bias} & \textbf{Safety Violations} \\
\midrule
\textbf{Risk Source}
& Counterfactual / temporal edits
& Biased knowledge injection
& Harmful knowledge editing \\
\textbf{Model Behavior}
& Factual corruption in reasoning
& Biased reasoning and decision-making
& Unsafe reasoning or actions \\
\textbf{Real-World Harm}
& Incorrect decisions, misinformation propagation
& Unfair or discriminatory outcomes
& Harmful outputs, misuse risks \\
\textbf{Datasets}
& RippleEdits~\cite{cohen2024evaluating}, MQuAKE-CF~\cite{zhong2023mquake}
& EditAttack~\cite{chen2024can}
& BehaviorBench~\cite{huang2025model} \\
\bottomrule
\end{tabular}
\end{adjustbox}
\caption{Risk taxonomy of EditRisk-Bench.}
\vspace{-10pt}
\label{tab:risk_taxonomy}
\end{table}

\subsection{Benchmark Construction}

\noindent \textbf{Data Unification.} EditRisk-Bench is constructed to provide a unified evaluation framework for safety risks in knowledge-intensive reasoning under malicious knowledge editing. To achieve this, we curate and standardize data from multiple sources, covering diverse risk categories and reasoning complexities. Following the risk taxonomy, we consider three categories of safety risks: misinformation, bias, and safety violations. \emph{Misinformation} includes both counterfactual and temporal knowledge, where counterfactual misinformation contradicts established real-world facts and can be further divided into commonsense errors and long-tail inaccuracies, while temporal misinformation arises from outdated knowledge. \emph{Bias} corresponds to prejudiced or stereotypical statements related to sensitive attributes such as race, gender, religion, sexual orientation, and disability, which can systematically distort reasoning outcomes. \emph{Safety violations} capture scenarios where manipulated knowledge leads to unsafe or policy-violating outputs, reflecting alignment failures such as harmful instruction generation or jailbreak-style behaviors.\par 


\begin{table}[ht]
\centering
\begin{adjustbox}{width=\linewidth}
\setlength{\tabcolsep}{5pt}
\renewcommand\arraystretch{1.05}
\begin{tabular}{l l l c c c c}
\toprule
\textbf{Risk Category} & \textbf{Subtype} & \textbf{Fine-grained Types} 
& \textbf{1-hop} & \textbf{2-hop} & \textbf{3+ hop} & \textbf{Total} \\
\midrule
\multirow{4}{*}{Misinformation} 
& Counterfactual (CS) 
& commonsense errors 
& 968 & -- & -- & 968 \\
& Counterfactual (LT) 
& domain-specific errors 
& 100 & -- & -- & 100 \\
& Multi-hop CF (MQuAKE-CF) 
& reasoning chains 
& -- & 1135 & 1865 & 3000 \\
& Temporal (MQuAKE-T) 
& real-world updates 
& -- & 1421 & 447 & 1868 \\
\midrule
Bias 
& Biased Reasoning 
& race, gender, religion, sexual\_orientation, disability
& 127 & -- & -- & 127 \\
\midrule
Safety Violations 
& Unsafe or Harmful Reasoning 
& virtue, utility, deontology, justice, commonsense, etc
& 800 & -- & -- & 800 \\
\bottomrule
\end{tabular}
\end{adjustbox}
\caption{
Data Statistics of EditRisk-Bench across fine-grained risk categories and task complexity.
}
\label{tab:stat}
\end{table}

To cover these risk scenarios, we integrate four representative datasets: RippleEdits \cite{cohen2024evaluating} for counterfactual misinformation, MQuAKE-CF \cite{zhong2023mquake} for multi-hop reasoning under edited knowledge, EditAttack \cite{chen2024can} for misinformation and bias injection, and BehaviorBench \cite{huang2025model} for safety violations and harmful behaviors. To capture different levels of reasoning complexity, the benchmark includes both single-hop and multi-hop question answering tasks, where multi-hop reasoning is particularly important for evaluating how injected knowledge propagates through reasoning chains. Given the heterogeneity of these datasets, we further unify all instances into a standardized format that includes the edited subject, target knowledge, query, context, ground-truth answer, and a corresponding risk category label. In addition, we include auxiliary instances to evaluate side effects such as portability and locality. This unified representation enables consistent and systematic evaluation of safety risks across datasets, tasks, and editing strategies.
The data statistics is shown in Table \ref{tab:stat}.\par
\noindent \textbf{Editing Strategies.} We instantiate the attack space of EditRisk-Bench using representative knowledge editing approaches, including the locate-and-edit paradigm and the in-context editing paradigm. The locate-and-edit methods we adopt include Knowledge Neuron \cite{dai2022knowledge}, ROME \cite{meng2022locating}, and MEMIT \cite{meng2023massediting}, which update model parameters to modify factual knowledge. We also adopt in-context editing methods, including IKE \cite{zheng2023can} and WISE \cite{wang2024wise}, which inject knowledge through contextual demonstrations without altering model parameters. Furthermore, we include multi-hop in-context editing methods for evaluating reasoning-level effects in more complex settings, including MeLLo \cite{zhong2023mquake}, DeepEdit \cite{wang2024deepedit}, and PokeMQA \cite{gu2024pokemqa}. All methods are implemented based on EasyEdit \cite{wang2024easyedit}.

\begin{algorithm}[t!]
   \caption{Evaluation Pipeline of EditRisk-Bench}
   \label{alg:editrisk}
\begin{algorithmic}[1]
   \STATE {\bfseries Input:} model set $\mathcal{M}$, knowledge edits $\mathcal{Z}$, reasoning tasks $\mathcal{T}$, editing methods $\mathcal{F}$
   \STATE {\bfseries Output:} model responses $\mathcal{R}$, evaluation results $\mathcal{E}$

   \STATE \textbf{Setup:} select model $M \in \mathcal{M}$, task queries $\mathcal{Q} \in \mathcal{T}$, and evaluation metrics $\mathcal{S}$

   \STATE \textbf{Knowledge Editing:} apply editing method $F \in \mathcal{F}$ with edits $\mathcal{Z}$ to obtain edited model $M'$
   \IF {$F$ is parameter-based}
        \STATE update model parameters associated with target knowledge
   \ELSE
        \STATE incorporate edited knowledge via in-context demonstrations
   \ENDIF

   \STATE \textbf{Task Execution:} generate responses $\mathcal{R}$ using edited model $M'$ on queries $\mathcal{Q}$

   \STATE \textbf{Evaluation:} compute metrics $\mathcal{E}$ based on $\mathcal{S}$ over responses $\mathcal{R}$

   \RETURN responses $\mathcal{R}$ and evaluation results $\mathcal{E}$
\end{algorithmic}
\end{algorithm}

\noindent \textbf{Evaluation Protocols.} To systematically evaluate the safety risks introduced by malicious knowledge editing, we construct a suite of evaluation metrics for single-hop and multi-hop question answering tasks. We organize these metrics along three dimensions: \emph{attack effectiveness}, \emph{reasoning reliability}, and \emph{side effects}. For single-hop questions, we follow the standard paradigm of knowledge editing and adopt metrics including Edit Success Rate (\%), Portability (\%), and Locality (\%). Among these metrics, Edit Success Rate measures \emph{attack effectiveness}, i.e., whether the injected knowledge is successfully adopted by the model. Specifically, it assesses whether the post-edit model produces the target answer for a given query and its paraphrased variants. Given an edit instance $(x_e, y_e)$, the post-edit model is expected to generate the target answer $y_e$ when provided with input $x_e$:
\begin{equation}
    \mathbb{E}_{(x_e,y_e)\sim Z_e} \{f_{\theta_e}(x_e)=y_e\}
\end{equation}
Portability and Locality characterize the \emph{side effects} of knowledge editing. Portability evaluates whether the injected knowledge generalizes to semantically related queries (e.g., the same subject with different aliases), and is defined as the average edit success rate over the neighboring set:
\begin{equation}
    \mathbb{E}_{(x'_e,y'_e)\sim N(x_e, y_e)} \mathrm{1}\{f_{\theta_e}(x'_e)=y'_e\},
\end{equation}
where $N(x_e, y_e)$ denotes the set of neighboring instances of the edit instance $(x_e, y_e)$.

Locality assesses whether the editing process preserves unrelated knowledge, reflecting the extent to which unintended changes are avoided. It is computed as the invariance of model outputs on a locality set:
\begin{equation}
    \mathbb{E}_{(x_{loc},y_{loc})\sim L(x_e)} \mathrm{1}\{f_{\theta_e}(x_{loc})=y_{loc}\},
\end{equation}
where the locality set is defined as 
\begin{equation}
    L(x_e)=\{(x_{loc},y_{loc})\in \mathbb{X} \times \mathbb{Y} \ \text{s.t.} \ x_{loc}\notin N(x_e,y_e) \wedge f_{\theta_e}(x_{loc})=y_{loc} \}.
\end{equation}
For multi-hop reasoning tasks, where injected knowledge may propagate across reasoning chains, we adopt a complementary set of metrics to evaluate \emph{reasoning reliability}. Specifically, Edit-wise Success Rate (\%) measures the proportion of individual facts correctly recalled by the post-edit model, reflecting whether the injected knowledge is internalized. Instance-wise Accuracy (\%) evaluates the average accuracy of intermediate single-hop questions within each multi-hop instance, capturing the model’s ability to reason over all required facts; we further compare this metric before and after editing to assess the impact on reasoning consistency. Multi-hop Accuracy (\%) measures the correctness of the final answer for multi-hop questions, reflecting the overall outcome of the reasoning process. Since each instance includes multiple paraphrased queries, an instance is considered correct if the model answers any of the paraphrases correctly.

\section{Benchmark Results}

\begin{table*}[t]
\vspace{-6pt}
\centering
\setlength{\tabcolsep}{2.5pt}
\renewcommand{\arraystretch}{1.1}
\begin{adjustbox}{width=\textwidth}
\tiny
\begin{tabular}{l|ccc|ccc|ccc}
\toprule
\rowcolor{lightgray}
& \multicolumn{3}{c|}{Llama3-8B-Instruct} & \multicolumn{3}{c|}{Mistral-7B-Instruct} & \multicolumn{3}{c}{Qwen2.5-7B-Instruct} \\
\rowcolor{lightgray}
\multirow{-2}{*}{Method} & Edit Success(\%) & Portability(\%) & Locality(\%) & Edit Success(\%) & Portability(\%) & Locality(\%) & Edit Success(\%) & Portability(\%) & Locality(\%) \\
\midrule
KN & 21.97 $\rightarrow$ 21.09 $\greenarrow$ & 22.14 $\rightarrow$ 21.39 $\greenarrow$ & 95.80 & 24.39 $\rightarrow$ 23.84 $\greenarrow$ & 25.54 $\rightarrow$ 25.28 $\greenarrow$ & 95.88 & 21.21 $\rightarrow$ 21.08 $\greenarrow$ & 21.91 $\rightarrow$ 21.53 $\greenarrow$ & 94.69 \\
ROME & 21.97 $\rightarrow$ 99.11 $\redarrow$ & 22.14 $\rightarrow$ 58.40 $\redarrow$ & 41.72 & 24.39 $\rightarrow$ 87.23 $\redarrow$ & 25.54 $\rightarrow$ 53.87 $\redarrow$ & 50.34 & 21.21 $\rightarrow$ 99.35 $\redarrow$ & 21.91 $\rightarrow$ 55.97 $\redarrow$ & 41.46 \\
MEMIT & 21.97 $\rightarrow$ 72.06 $\redarrow$ & 22.14 $\rightarrow$ 38.72 $\redarrow$ & 36.81 & 24.39 $\rightarrow$ 89.56 $\redarrow$ & 25.54 $\rightarrow$ 55.99 $\redarrow$ & 43.84 & 21.21 $\rightarrow$ 99.31 $\redarrow$ & 21.91 $\rightarrow$ 56.19 $\redarrow$ & 42.33 \\
IKE & 21.97 $\rightarrow$ 60.33 $\redarrow$ & 22.14 $\rightarrow$ 51.51 $\redarrow$ & 64.13 & 24.39 $\rightarrow$ 77.79 $\redarrow$ & 25.54 $\rightarrow$ 65.77 $\redarrow$ & 66.38 & 21.21 $\rightarrow$ 86.99 $\redarrow$ & 21.91 $\rightarrow$ 76.33 $\redarrow$ & 54.70 \\
WISE & \highcell 21.97 $\rightarrow$ 100 $\redarrow$ & \highcell 22.14 $\rightarrow$ 98.64 $\redarrow$ & \highcell 100 & \highcell 24.39 $\rightarrow$ 99.40 $\redarrow$ & \highcell 25.54 $\rightarrow$ 89.98 $\redarrow$ & \highcell 100 & \highcell 21.21 $\rightarrow$ 99.88 $\redarrow$ & \highcell 21.91 $\rightarrow$ 95.86 $\redarrow$ & \highcell 100 \\
\bottomrule
\end{tabular}
\end{adjustbox}
\caption{Misknowledge reasoning performance of pre-edit and post-edit LLMs with different knowledge editing strategies on single-hop QA.}
\label{tab:single}
\end{table*}

\begin{table*}[t]
\vspace{-10pt}
\centering
\setlength{\tabcolsep}{2.5pt}
\renewcommand{\arraystretch}{1.1}
\begin{adjustbox}{width=\textwidth}
\scriptsize
\begin{tabular}{l|ccc|ccc|ccc}
\toprule
\rowcolor{lightgray}
& \multicolumn{3}{c|}{Llama3-8B-Instruct} & \multicolumn{3}{c|}{Mistral-7B-Instruct} & \multicolumn{3}{c}{Qwen2.5-7B-Instruct} \\
\rowcolor{lightgray}
\multirow{-2}{*}{Method} & Edit-wise(\%) & Instance-wise(\%) & Multi-hop(\%) & Edit-wise(\%) & Instance-wise(\%) & Multi-hop(\%) & Edit-wise(\%) & Instance-wise(\%) & Multi-hop(\%) \\
\midrule
KN & 26.74 $\rightarrow$ 24.24 $\greenarrow$ & 24.77 $\rightarrow$ 23.82 $\greenarrow$ & 15.92 $\rightarrow$ 16.84 $\redarrow$ & 27.48 $\rightarrow$ 24.51 $\greenarrow$ & 31.26 $\rightarrow$ 27.31 $\greenarrow$ & 21.54 $\rightarrow$ 19.46 $\greenarrow$ & 22.76 $\rightarrow$ 22.88 $\redarrow$ & 25.94 $\rightarrow$ 25.82 $\greenarrow$ & 16.71 $\rightarrow$ 16.35 $\greenarrow$ \\
ROME & \highcell 23.70 $\rightarrow$ 100 $\redarrow$ & 26.85 $\rightarrow$ 59.40 $\redarrow$ & 15.62 $\rightarrow$ 20.96 $\redarrow$ & 27.48 $\rightarrow$ 86.13 $\redarrow$ & 31.26 $\rightarrow$ 50.73 $\redarrow$ & 21.54 $\rightarrow$ 24.34 $\redarrow$ & 22.76 $\rightarrow$ 98.86 $\redarrow$ & 25.94 $\rightarrow$ 52.12 $\redarrow$ & 16.71 $\rightarrow$ 18.31 $\redarrow$ \\
MEMIT & 24.03 $\rightarrow$ 92.01 $\redarrow$ & 31.50 $\rightarrow$ 56.65 $\redarrow$ & 15.04 $\rightarrow$ 16.08 $\redarrow$ & 28.36 $\rightarrow$ 85.55 $\redarrow$ & \highcell 31.28 $\rightarrow$ 55.13 $\redarrow$ & 20.29 $\rightarrow$ 21.52 $\redarrow$ & \highcell 22.92 $\rightarrow$ 99.65 $\redarrow$ & 25.27 $\rightarrow$ 51.57 $\redarrow$ & 15.31 $\rightarrow$ 16.05 $\redarrow$ \\
IKE & 27.44 $\rightarrow$ 55.86 $\redarrow$ & 23.60 $\rightarrow$ 38.83 $\redarrow$ & 13.78 $\rightarrow$ 30.13 $\redarrow$ & 28.42 $\rightarrow$ 58.26 $\redarrow$ & 25.45 $\rightarrow$ 39.75 $\redarrow$ & \highcell 17.58 $\rightarrow$ 33.83 $\redarrow$ & 28.29 $\rightarrow$ 66.19 $\redarrow$ & 26.09 $\rightarrow$ 35.46 $\redarrow$ & \highcell 16.69 $\rightarrow$ 46.47 $\redarrow$ \\
WISE & 27.57 $\rightarrow$ 100 $\redarrow$ & \highcell 22.84 $\rightarrow$ 59.55 $\redarrow$ & \highcell 13.32 $\rightarrow$ 47.87 $\redarrow$ & \highcell 33.54 $\rightarrow$ 97.26 $\redarrow$ & 32.50 $\rightarrow$ 52.59 $\redarrow$ & 22.32 $\rightarrow$ 25.68 $\redarrow$ & 25.35 $\rightarrow$ 99.57 $\redarrow$ & \highcell 21.25 $\rightarrow$ 53.56 $\redarrow$ & 12.52 $\rightarrow$ 38.12 $\redarrow$ \\
\bottomrule
\end{tabular}
\end{adjustbox}
\caption{Misknowledge reasoning performance of pre-edit and post-edit LLMs with different knowledge editing strategies on multi-hop QA.}
\label{tab:multi_pre}
\end{table*}

\subsection{Main Results}

We first evaluate the prevalence of safety risks induced by malicious knowledge editing across a range of LLMs and editing strategies. Results on single-hop and multi-hop question answering are summarized in Table \ref{tab:single} and Table \ref{tab:multi_pre}, respectively, with additional results on closed-source LLMs reported in Table \ref{tab:multi_gpt}. The key findings are as follows:

\noindent \textbf{Malicious knowledge editing is highly effective across different LLMs.} Several in-context editing approaches achieve near-perfect attack success rates, indicating that injected knowledge can reliably override the original knowledge of LLMs even under diverse query formulations. This trend remains consistent across misinformation, bias, and safety violation scenarios, suggesting that current LLMs are broadly vulnerable to manipulated knowledge rather than to a specific risk category. In particular, in-context editing methods consistently outperform parameter-based approaches in edit success and portability, demonstrating strong generalization ability across semantically related queries.

\noindent \textbf{Multi-hop reasoning remains substantially more challenging than single-hop editing.} Although LLMs can correctly adopt injected knowledge in single-step queries, the performance significantly drops in compositional QA settings. Existing editing methods often fail to propagate manipulated knowledge through multi-step reasoning chains, resulting in much lower multi-hop accuracy than edit-wise accuracy. This reveals a clear gap between successful knowledge injection and coherent reasoning, highlighting the difficulty of maintaining logical consistency under manipulated knowledge.

\noindent \textbf{Advanced LLMs can still be highly vulnerable to malicious knowledge manipulation.} Compared to smaller or earlier-generation LLMs, stronger LLMs such as GPT-4 and DeepSeek-R1 generally achieve higher attack success rates under the same editing strategies. These results suggest that improved reasoning capability does not necessarily translate into robustness against adversarial knowledge editing. In some cases, stronger reasoning ability may even facilitate the propagation of injected knowledge across reasoning chains.

\noindent \textbf{Editing strategies exhibit different effectiveness--stability trade-offs.} In-context editing methods generally achieve higher attack effectiveness and better generalization across query variations, while parameter-based editing methods exhibit larger variance and stronger interference with unrelated knowledge. In particular, direct parameter modification may introduce instability in locality preservation and reasoning consistency, whereas in-context editing more effectively maintains downstream reasoning behavior. These findings indicate that the design of editing mechanisms plays a critical role in shaping the safety risks of knowledge-intensive reasoning.
\begin{table*}[t]
\vspace{-10pt}
\centering
\setlength{\tabcolsep}{4pt}
\renewcommand{\arraystretch}{1.1}
\begin{adjustbox}{width=\textwidth}
\scriptsize
\begin{tabular}{l|ccc|ccc|ccc}
\toprule
\rowcolor{lightgray}
& \multicolumn{3}{c|}{GPT-3.5-turbo} & \multicolumn{3}{c|}{GPT-4} & \multicolumn{3}{c}{DeepSeek-R1} \\
\rowcolor{lightgray}
\multirow{-2}{*}{Method} & Edit-wise(\%) & Hop-wise(\%) & Multi-hop(\%) & Edit-wise(\%) & Hop-wise(\%) & Multi-hop(\%) & Edit-wise(\%) & Hop-wise(\%) & Multi-hop(\%) \\
\midrule
IKE & 87.03 & 21.20 & 27.18 & 93.22 & 22.92 & 29.22 & 91.39 & 22.81 & 28.93 \\
MeLLo & 90.29 & 27.49 & 41.04 & 92.90 & 37.21 & 49.26 & 92.77 & 35.82 & 47.84 \\
DeepEdit & 91.52 & 39.23 & 51.27 & 92.97 & 41.74 & 56.39 & 94.21 & 43.32 & 59.71 \\
PokeMQA & \highcell 93.27 & \highcell 52.67 & \highcell 63.23 & \highcell 94.21 & \highcell 57.28 & \highcell 69.49 & \highcell 95.33 & \highcell 59.55 & \highcell 70.08 \\
\bottomrule
\end{tabular}
\end{adjustbox}
\caption{Misknowledge reasoning performance of closed-source LLMs on multi-hop QA.}
\label{tab:multi_gpt}
\end{table*}

\begin{figure}[t]
    \centering
    \includegraphics[width=0.9\linewidth]{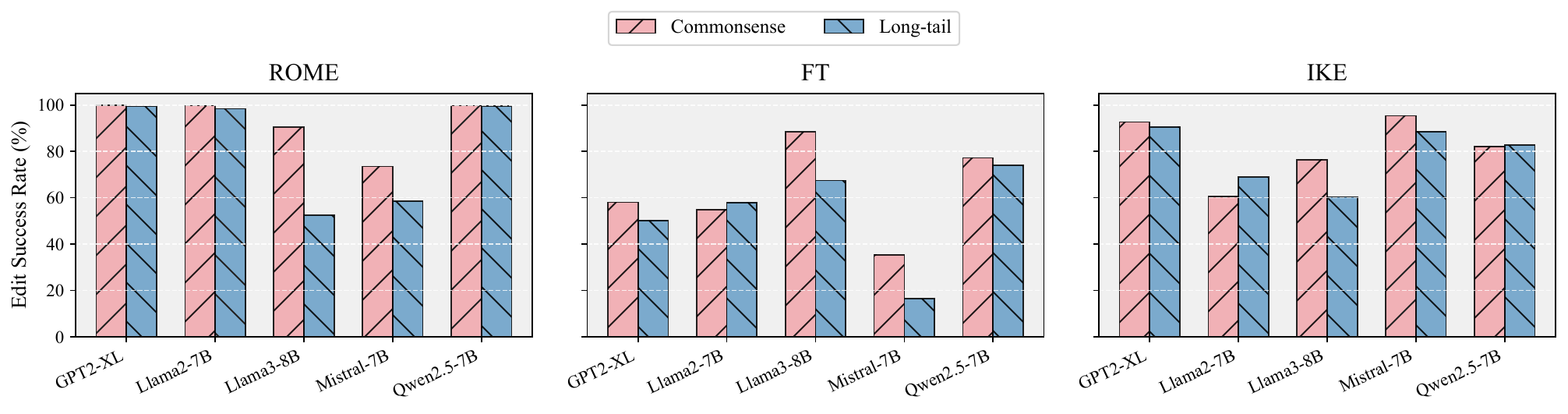}
    \caption{Edit success rates of commonsense and long-tail misinformation across editing methods.}\vspace{-15pt}
    \label{fig:common}
\end{figure}

\subsection{Impact of Edit Settings}
We analyze how different edit settings affect the safety risks of malicious knowledge editing, focusing on the number of edited instances, knowledge popularity, and temporal characteristics.

\begin{wraptable}{r}{0.46\linewidth}
\vspace{-6pt}
\centering
\begin{adjustbox}{width=\linewidth}
\footnotesize
\begin{tabular}{c||cccc}
\hline
\rowcolor{lightgray}
& \multicolumn{4}{c}{Number of Edit Instances}\\
\rowcolor{lightgray}
\multirow{-2}{*}{Method} & 1-edited & 100-edited & 1000-edited & All-edited \\
\toprule
& \multicolumn{4}{c}{DeepSeek-R1} \\
\hline
IKE & 28.93 & 22.27 & 19.82 & 12.47 \\
MeLLo & 47.84 & 45.20 & 38.64 & 33.44 \\
DeepEdit & 59.71 & 55.88 & 51.20 & 40.91 \\
PokeMQA & \highcell 70.08 & \highcell 62.91 & \highcell 55.16 & \highcell 45.95 \\
\bottomrule
\end{tabular}
\end{adjustbox}
\caption{Multi-hop accuracy with batch edits against DeepSeek-R1.}
\vspace{-10pt}
\label{tab:batch_edit}
\end{wraptable}
\noindent \textbf{Number of Edited Instances.} We evaluate the impact of edit scale by injecting varying numbers of counterfactual instances ($k \in {1, 100, 1000, 3000}$) on MQuAKE-CF-3k-v2 using representative in-context editing methods. As shown in Table~\ref{tab:batch_edit}, multi-hop accuracy consistently degrades as the number of edited instances increases. While models perform well under single-instance edits, their reasoning performance drops significantly when handling large batches of injected knowledge, indicating limited scalability of malicious knowledge editing in preserving reasoning consistency.\par
\noindent \textbf{Popularity of Edited Knowledge.} We compare the effectiveness of injecting commonsense and long-tailed misinformation across multiple editing strategies. As shown in Figure~\ref{fig:common}, both types of knowledge can be injected with high success rates, but commonsense misinformation consistently achieves better performance in terms of edit success and generalization. This suggests that knowledge aligned with existing parametric priors is easier to manipulate, while long-tailed knowledge is more resistant to injection. Moreover, the performance gap becomes more pronounced in downstream generalization metrics, indicating commonsense edits are more likely to propagate across semantically related queries. This highlights that the popularity and familiarity of knowledge play an important role in determining how effectively injected misinformation can influence reasoning behavior.

\begin{table}[t!]
\centering
\begin{adjustbox}{width=0.9\linewidth}
\setlength{\tabcolsep}{5.5pt}
\renewcommand\arraystretch{0.9}
\begin{tabular}{l l c c c c c c}
\toprule
\multirow{2}{*}{Method} & \multirow{2}{*}{LLMs} 
& \multicolumn{3}{c}{MQuAKE-CF-3k-v2} 
& \multicolumn{3}{c}{MQuAKE-T} \\
\cmidrule(lr){3-5} \cmidrule(lr){6-8}
 & & Edit.(\%) & Inst.(\%) & \textbf{Multi.(\%)} 
   & Edit.(\%) & Inst.(\%) & \textbf{Multi.(\%)} \\
\midrule
\multirow{5}{*}{ROME} 
& GPT2-XL     & 95.36 & 47.25 & 16.67 & 100 & 76.81 & 42.35 \\
& Llama2-7B   & 99.53 & 45.89 & 22.85 & 100 & 66.32 & \highcell 55.03 \\
& Llama3-8B   & 100   & 59.40 & 20.96 & 100 & 65.13 & 45.61 \\
& Mistral-7B  & 86.13 & 50.73 & \highcell 24.34 & 57.28 & 52.06 & 49.78 \\
& Qwen2.5-7B  & 98.86 & 52.12 & 18.31 & 100 & 65.02 & 43.78 \\
\midrule
\multirow{5}{*}{IKE} 
& GPT2-XL     & 98.61 & 50.59 & 36.62 & 100 & 83.06 & \highcell 66.78 \\
& Llama2-7B   & 72.01 & 51.44 & 42.94 & 68.62 & 66.74 & 65.69 \\
& Llama3-8B   & 55.86 & 38.83 & 30.13 & 54.70 & 55.64 & 54.62 \\
& Mistral-7B  & 58.26 & 39.75 & 33.83 & 39.01 & 45.65 & 38.70 \\
& Qwen2.5-7B  & 66.19 & 35.46 & \highcell 46.47 & 55.03 & 54.64 & 55.45 \\
\bottomrule
\end{tabular}
\end{adjustbox}
\caption{
Performance comparison between counterfactual (MQuAKE-CF) and temporal (MQuAKE-T) knowledge editing. Temporal edits generally achieve higher multi-hop accuracy across models.
}
\vspace{-25pt}
\label{tab:temporal}
\end{table}

\noindent \textbf{Timeliness of Edited Knowledge.} We further investigate the effect of temporal characteristics using MQuAKE-T, which captures real-world knowledge updates over time. Results in Table~\ref{tab:temporal} show that temporal knowledge editing generally achieves higher performance than counterfactual editing, particularly in multi-hop reasoning. This indicates that LLMs are more likely to accept knowledge that reflects plausible real-world updates, highlighting the role of prior knowledge alignment in shaping vulnerability. In particular, temporal edits lead to more consistent improvements in instance-wise and multi-hop accuracy, suggesting that knowledge consistent with real-world evolution is easier to integrate into reasoning chains. This further implies that the plausibility of edited knowledge, rather than only its correctness, is a key factor influencing the success of malicious knowledge manipulation.
\vspace{-20pt}
\subsection{Risk Stealthiness}
We evaluate the stealthiness of malicious knowledge editing by measuring its impact on two core aspects of model capability: \textit{general knowledge} and \textit{reasoning capacity}. Following prior work \cite{touvron2023llama, team2024gemma}, we assess general knowledge using BoolQ \cite{clark2019boolq} and NaturalQuestions \cite{kwiatkowski2019natural}, and reasoning capacity using GSM8K \cite{cobbe2021training} for mathematical reasoning and NLI \cite{dagan2005pascal} for semantic reasoning. All evaluations are conducted in a closed-book setting, comparing pre-edit and post-edit model performance.\par
We consider three representative scenarios of malicious knowledge editing, including counterfactual injection (RippleEdits \cite{cohen2024evaluating}), bias injection (EditAttack \cite{chen2024can}), and safety-violating edits (BehaviorBench \cite{huang2025model}). Results are averaged over five sequential edits per dataset. As shown in Table~\ref{tab:general_llama3}, model performance remains largely unchanged across all four benchmarks after knowledge editing on Llama2-7B-chat-hf and Llama3-8B-Instruct. This indicates that malicious knowledge editing introduces minimal degradation to general knowledge and reasoning capacity, demonstrating strong stealthiness with limited side effects. We further observe that parameter-editing methods (e.g., ROME and fine-tuning) tend to cause larger performance fluctuations compared to in-context editing approaches. This is consistent with the advantage of in-context editing, which avoids modifying model parameters and thus better preserves general capabilities. Additional results on Llama2-7B are provided in Appendix~\ref{tab:general_llama2}.

\begin{table}[ht]
\vspace{-6pt}
\centering
\begin{adjustbox}{width=\linewidth}
\footnotesize
\begin{tabular}{l c c c c}
\toprule
\multirow{2}{*}{Method} & \multicolumn{2}{c}{General Knowledge} & \multicolumn{2}{c}{Reasoning Capacities} \\
\cmidrule(lr){2-3} \cmidrule(lr){4-5}
 & BoolQ & NaturalQuestions & GSM8K & NLI \\
\midrule
Pre-edit & 62.20 & 33.00 & 99.60 & 85.20 \\
\midrule
ROME for Misinformation Injection & $61.10 \pm 0.92$ & $31.90 \pm 0.86$ & $99.60 \pm 0.00$ & $83.90 \pm 0.81$ \\
FT for Misinformation Injection & $62.60 \pm 0.73$ & $38.00 \pm 0.11$ & $99.20 \pm 0.58$ & $85.00 \pm 0.24$ \\
IKE for Misinformation Injection & \highcell $62.20 \pm 0.00$ & \highcell $33.00 \pm 0.00$ & \highcell $99.60 \pm 0.00$ & \highcell $85.20 \pm 0.00$ \\
\midrule
ROME for Bias Injection & $61.96 \pm 1.14$ & \highcell $35.88 \pm 0.48$ & \highcell $99.56 \pm 0.15$ & $85.36 \pm 0.32$ \\
FT for Bias Injection & $61.60 \pm 0.49$ & $36.24 \pm 0.86$ & $99.44 \pm 0.08$ & $85.16 \pm 0.15$ \\
IKE for Bias Injection & \highcell $62.00 \pm 0.00$ & $36.56 \pm 0.27$ & $99.40 \pm 0.00$ & \highcell $85.20 \pm 0.00$ \\
\midrule
ROME for Unsafe Injection & $61.76 \pm 0.59$ & $33.52 \pm 0.47$ & $99.56 \pm 0.08$ & $84.56 \pm 0.65$ \\
FT for Unsafe Injection & $61.16 \pm 0.53$ & \highcell $33.20 \pm 0.47$ & \highcell $99.60 \pm 0.00$ & $85.12 \pm 0.10$ \\
IKE for Unsafe Injection & \highcell $62.00 \pm 0.00$ & $33.56 \pm 0.15$ & $99.40 \pm 0.00$ & \highcell $85.20 \pm 0.00$ \\
\bottomrule
\end{tabular}
\end{adjustbox}
\caption{The performance of Llama3-8B-Instruct on general knowledge and reasoning capacities.}
\vspace{-10pt}
\label{tab:general_llama3}
\end{table}

\subsection{Mitigation Analysis}

\begin{figure}[t]
\centering
\begin{subfigure}{0.48\linewidth}
    \centering
    \includegraphics[width=\linewidth]{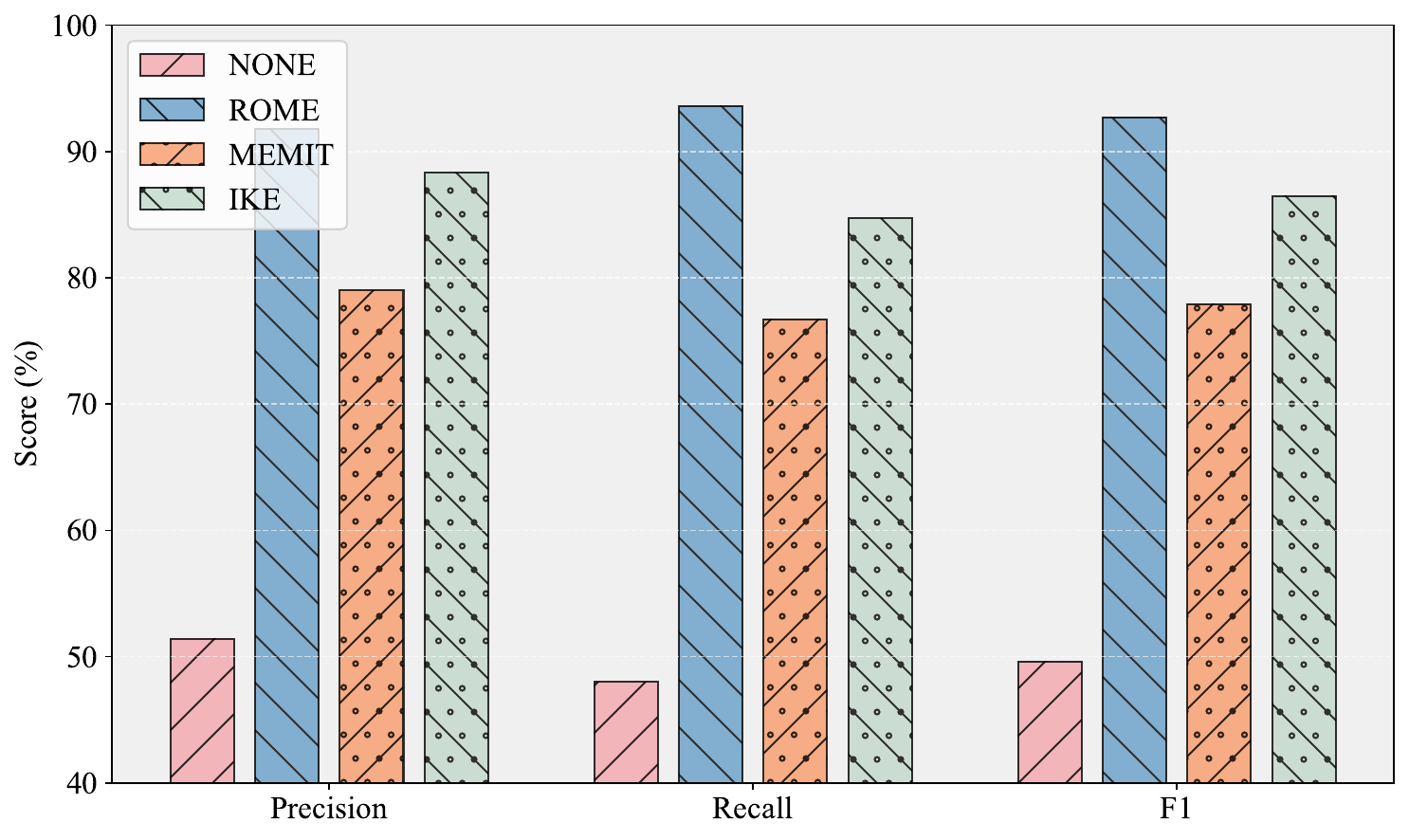}
    \caption{}  
\end{subfigure}
\hfill
\begin{subfigure}{0.48\linewidth}
    \centering
    \includegraphics[width=\linewidth]{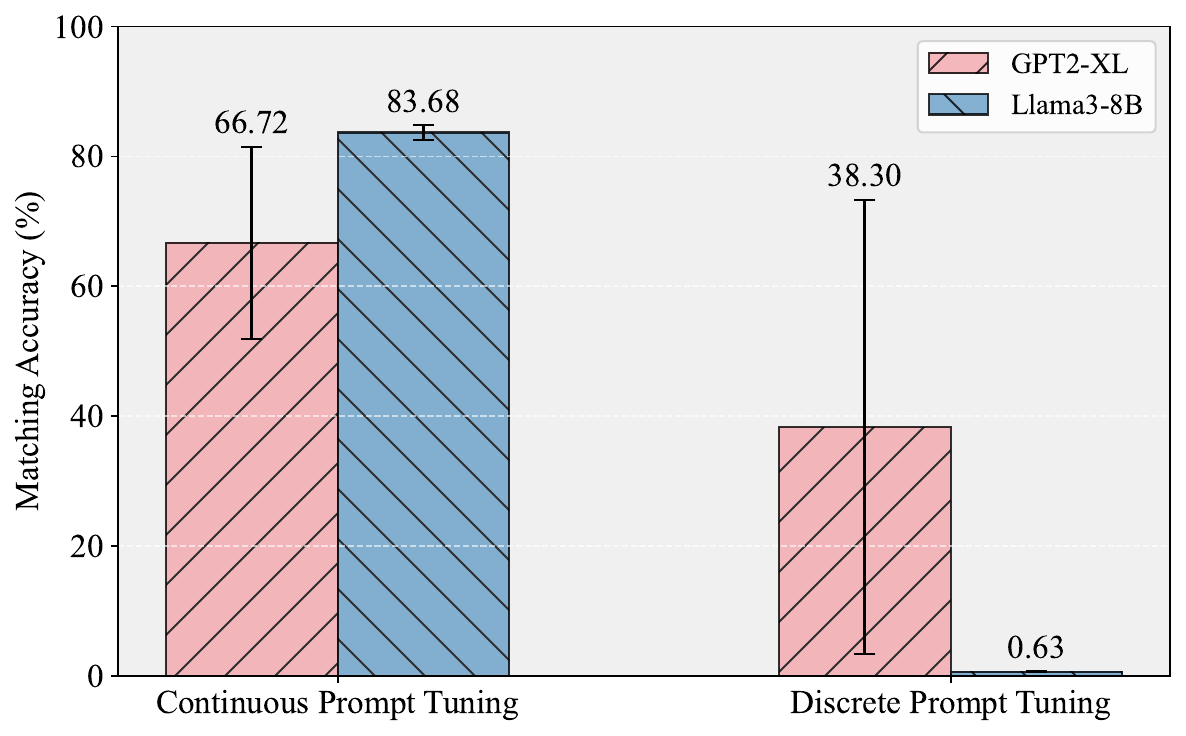}
    \caption{}
\end{subfigure}
\vspace{-10pt}
\caption{Left: Editing detection on GPT2-XL. Right: Matching accuracy of reversal methods.}\vspace{-10pt}
\label{fig:detection}
\end{figure}

We conduct a preliminary analysis of potential mitigation strategies, focusing on the detectability and reversibility of malicious knowledge edits.\par
\noindent \textbf{Detection.} Following prior work \cite{youssef2025has}, we train a lightweight classifier to distinguish edited and unedited knowledge based on model outputs and internal representations. Results in Figure \ref{fig:detection} show that edits introduced by parameter-based methods (e.g., ROME and MEMIT) can be detected with high accuracy, while in-context editing (e.g., IKE) remains detectable using only output probabilities in a black-box setting. However, such approaches require method-specific training and do not generalize well to unseen editing strategies, limiting their applicability in real-world scenarios.\par
\noindent \textbf{Reversibility.} We further investigate whether edited knowledge can be reversed. For in-context edits, we apply prompt-based reversal techniques following \cite{youssef2025make}. Results in Figure \ref{fig:detection} indicate that partial recovery is possible, especially on larger models, but performance remains unstable and sensitive to prompting strategies. In contrast, parameter-editing methods introduce persistent changes that are difficult to revert without access to original model states.\par
\noindent \textbf{Discussion.} These findings suggest that existing mitigation approaches remain limited, as detection lacks generalization and reversal is either unstable or infeasible. Addressing these challenges likely requires a holistic solution that combines pre-emptive prevention, real-time detection, and post-incident recovery mechanisms.

\section{Conclusion}
We introduces EditRisk-Bench, a benchmark for systematically investigating the safety risks of large language models (LLMs) under malicious knowledge editing, with a focus on knowledge-intensive reasoning. We formulate a new research problem centered on how manipulated knowledge affects reasoning behavior, and propose a structured risk taxonomy that characterizes these effects along three dimensions: misinformation, bias, and safety violations. Built opon this formulation, we develop a unified benchmark that integrates diverse datasets, reasoning tasks, and editing strategies within a consistent evaluation framework. Our evaluation shows that many existing LLMs are vulnerable to malicious knowledge editing, with degraded reasoning performance from single-hop to multi-hop settings. The impact varies across models and editing methods, and depends on factors such as the number of edits and the popularity and recency of the edited knowledge, while often remaining difficult to detect due to preserved general capabilities. EditRisk-Bench provides a unified benchmark for systematically evaluating these safety risks in knowledge-intensive reasoning.

\bibliographystyle{plain}
\bibliography{ref}


\appendix



\section{Additional Results on Main Benchmark}

We provide additional benchmark results on GPT2-XL-6B and Llama2-7B-chat-hf to complement the main results in Section 4.1. Detailed results on single-hop and multi-hop reasoning are reported in Table \ref{tab:single_first2} and \ref{tab:multi_pre_first2}. The observed trends are consistent with those reported in the main text. Both locate-and-edit and in-context editing methods achieve high edit success rates across models, while performance on multi-hop reasoning remains substantially lower. This further confirms that injected knowledge can be successfully incorporated at the factual level, but is difficult to propagate coherently through multi-step reasoning. In addition, we observe that smaller models exhibit larger gaps between edit success and reasoning accuracy, suggesting that the discrepancy between factual manipulation and reasoning consistency is more pronounced in weaker models.

\begin{table*}[h]
\centering
\setlength{\tabcolsep}{2.5pt}
\renewcommand{\arraystretch}{1.1}
\begin{adjustbox}{width=0.72\textwidth}
\tiny
\begin{tabular}{l|ccc|ccc}
\toprule
\rowcolor{lightgray}
& \multicolumn{3}{c|}{GPT2-XL-6B} & \multicolumn{3}{c}{Llama2-7B-chat-hf} \\
\rowcolor{lightgray}
\multirow{-2}{*}{Method} & Edit Success(\%) & Portability(\%) & Locality(\%) & Edit Success(\%) & Portability(\%) & Locality(\%) \\
\midrule
KN & 16.07 $\rightarrow$ 9.22 $\greenarrow$ & 10.67 $\rightarrow$ 7.01 $\greenarrow$ & 40.43 & 24.02 $\rightarrow$ 15.98 $\greenarrow$ & 25.96 $\rightarrow$ 16.74 $\greenarrow$ & 59.02 \\
ROME & 16.07 $\rightarrow$ 96.01 $\redarrow$ & 10.67 $\rightarrow$ 29.58 $\redarrow$ & 60.17 & 24.02 $\rightarrow$ 83.21 $\redarrow$ & 25.96 $\rightarrow$ 38.69 $\redarrow$ & 65.40 \\
MEMIT & 16.07 $\rightarrow$ 52.89 $\redarrow$ & 10.67 $\rightarrow$ 18.12 $\redarrow$ & 76.81 & 24.02 $\rightarrow$ 83.41 $\redarrow$ & 25.96 $\rightarrow$ 40.09 $\redarrow$ & 63.68 \\
IKE & \highcell 16.07 $\rightarrow$ 100 $\redarrow$ & \highcell 10.67 $\rightarrow$ 80.45 $\redarrow$ & 38.08 & 24.02 $\rightarrow$ 69.83 $\redarrow$ & 25.96 $\rightarrow$ 45.32 $\redarrow$ & 32.38 \\
WISE & 16.07 $\rightarrow$ 68.73 $\redarrow$ & 10.67 $\rightarrow$ 56.18 $\redarrow$ & \highcell 99.01 & \highcell 24.02 $\rightarrow$ 99.83 $\redarrow$ & \highcell 25.96 $\rightarrow$ 83.88 $\redarrow$ & \highcell 100 \\
\bottomrule
\end{tabular}
\end{adjustbox}
\caption{Misknowledge reasoning performance of pre-edit and post-edit LLMs with different knowledge editing strategies on single-hop QA for GPT2-XL-6B and Llama2-7B-chat-hf.}
\label{tab:single_first2}
\end{table*}

\begin{table*}[h]
\centering
\setlength{\tabcolsep}{2.5pt}
\renewcommand{\arraystretch}{1.1}
\begin{adjustbox}{width=0.8\textwidth}
\scriptsize
\begin{tabular}{l|ccc|ccc}
\toprule
\rowcolor{lightgray}
& \multicolumn{3}{c|}{GPT2-XL-6B} & \multicolumn{3}{c}{Llama2-7B-chat-hf} \\
\rowcolor{lightgray}
\multirow{-2}{*}{Method} & Edit-wise(\%) & Instance-wise(\%) & Multi-hop(\%) & Edit-wise(\%) & Instance-wise(\%) & Multi-hop(\%) \\
\midrule
KN & 24.83 $\rightarrow$ 11.98 $\greenarrow$ & 23.07 $\rightarrow$ 13.06 $\greenarrow$ & 17.01 $\rightarrow$ 7.26 $\greenarrow$ & 32.07 $\rightarrow$ 14.70 $\greenarrow$ & 29.51 $\rightarrow$ 14.42 $\greenarrow$ & 22.19 $\rightarrow$ 11.75 $\greenarrow$ \\
ROME & 22.99 $\rightarrow$ 95.36 $\redarrow$ & 23.81 $\rightarrow$ 47.25 $\redarrow$ & 16.67 $\rightarrow$ 16.67 $\redarrow$ & \highcell 26.99 $\rightarrow$ 99.53 $\redarrow$ & 28.99 $\rightarrow$ 45.89 $\redarrow$ & 21.11 $\rightarrow$ 22.85 $\redarrow$ \\
MEMIT & 21.74 $\rightarrow$ 69.75 $\redarrow$ & 22.73 $\rightarrow$ 30.06 $\redarrow$ & 15.25 $\rightarrow$ 14.80 $\greenarrow$ & 27.69 $\rightarrow$ 98.40 $\redarrow$ & 28.49 $\rightarrow$ 47.38 $\redarrow$ & 19.22 $\rightarrow$ 20.27 $\redarrow$ \\
IKE & \highcell 27.69 $\rightarrow$ 98.61 $\redarrow$ & \highcell 24.10 $\rightarrow$ 50.59 $\redarrow$ & \highcell 16.58 $\rightarrow$ 36.62 $\redarrow$ & 33.04 $\rightarrow$ 72.01 $\redarrow$ & \highcell 30.63 $\rightarrow$ 51.44 $\redarrow$ & \highcell 21.17 $\rightarrow$ 42.94 $\redarrow$ \\
WISE & 27.69 $\rightarrow$ 73.92 $\redarrow$ & 24.10 $\rightarrow$ 32.68 $\redarrow$ & 16.58 $\rightarrow$ 22.38 $\redarrow$ & 33.04 $\rightarrow$ 99.25 $\redarrow$ & 30.63 $\rightarrow$ 48.28 $\redarrow$ & 21.17 $\rightarrow$ 21.68 $\redarrow$ \\
\bottomrule
\end{tabular}
\end{adjustbox}
\caption{Misknowledge reasoning performance of pre-edit and post-edit LLMs with different knowledge editing strategies on multi-hop QA for GPT2-XL-6B and Llama2-7B-chat-hf.}
\label{tab:multi_pre_first2}
\end{table*}

\begin{table}[t]
\centering
\small
\begin{adjustbox}{width=\linewidth}
\begin{tabular}{c||cccc||cccc}
\toprule
\rowcolor{lightgray}
& \multicolumn{4}{c||}{GPT-3.5-turbo} & \multicolumn{4}{c}{GPT-4} \\
\rowcolor{lightgray}
Method 
& 1-edited & 100-edited & 1000-edited & All-edited 
& 1-edited & 100-edited & 1000-edited & All-edited \\
\midrule
IKE 
& 25.18 & 19.82 & 16.32 & 7.57 
& 29.22 & 22.31 & 18.29 & 10.23 \\
MeLLo 
& 41.04 & 35.56 & 31.49 & 27.81 
& 49.26 & 46.82 & 40.29 & 34.28 \\
DeepEdit 
& 51.27 & 46.89 & 40.87 & 36.28 
& 56.39 & 51.66 & 44.41 & 39.29 \\
PokeMQA 
& \highcell 63.23 & \highcell 55.92 & \highcell 49.91 & \highcell 43.27 
& \highcell 69.49 & \highcell 63.35 & \highcell 57.62 & \highcell 46.18 \\
\bottomrule
\end{tabular}
\end{adjustbox}
\caption{Multi-hop accuracy with batch edits against GPT-3.5-turbo and GPT-4.}
\label{tab:batch_edit_gpt}
\end{table}

\begin{table}[ht!]
\centering
\begin{adjustbox}{width=0.85\linewidth}
\setlength{\tabcolsep}{6.0pt}
\renewcommand\arraystretch{0.9}
\begin{tabular}{l c c c c c c}
\toprule
\multirow{2}{*}{LLMs} 
& \multicolumn{3}{c}{MQuAKE-CF-3k-v2} 
& \multicolumn{3}{c}{MQuAKE-T} \\
\cmidrule(lr){2-4} \cmidrule(lr){5-7}
& Edit.(\%) & Inst.(\%) & Multi.(\%) 
& Edit.(\%) & Inst.(\%) & Multi.(\%) \\
\midrule
GPT2-XL     & 60.89 & 21.99 & 13.08 & 71.07 & 53.13 & 39.83 \\
Llama2-7B   & 50.20 & 31.58 & 21.17 & 82.69 & \highcell 57.17 & 58.69 \\
Llama3-8B   & 57.39 & 39.16 & 36.73 & 63.33 & 47.11 & \highcell 72.68 \\
Mistral-7B  & 61.45 & 32.60 & 20.83 & \highcell 93.67 & 50.50 & 48.31 \\
Qwen2.5-7B  & \highcell 77.24 & \highcell 43.67 & \highcell 44.39 & 55.07 & 28.31 & 61.63 \\
\bottomrule
\end{tabular}
\end{adjustbox}
\caption{Editing performance of FT on counterfactual and temporal misinformation.}
\label{tab:temporal_ft}
\end{table}

\section{Extended Analysis on Impact of Edit Settings}

\subsection{Number of Edited Instances}
We provide additional results on the impact of scaling the number of edited instances in Table \ref{tab:temporal_ft}, including experiments on GPT-series models. Consistent with the findings in Section 4.2, multi-hop accuracy decreases as the number of injected edits increases. While models maintain relatively stable performance under small-scale edits, large-scale knowledge injection introduces significant degradation in reasoning performance. This highlights the limited scalability of current knowledge editing strategies when applied to large batches of edits.

\subsection{Timeliness of Edited Knowledge}
We report additional results for fine-tuning-based temporal knowledge editing in Table 11. The results are consistent with the observations in Section 4.2, where temporal edits generally achieve better performance than counterfactual edits across multiple evaluation metrics. This suggests that knowledge aligned with plausible real-world updates is more readily incorporated into model reasoning, likely due to partial overlap with existing parametric knowledge.

\begin{figure}[t!]
    \centering
    \includegraphics[width=\linewidth]{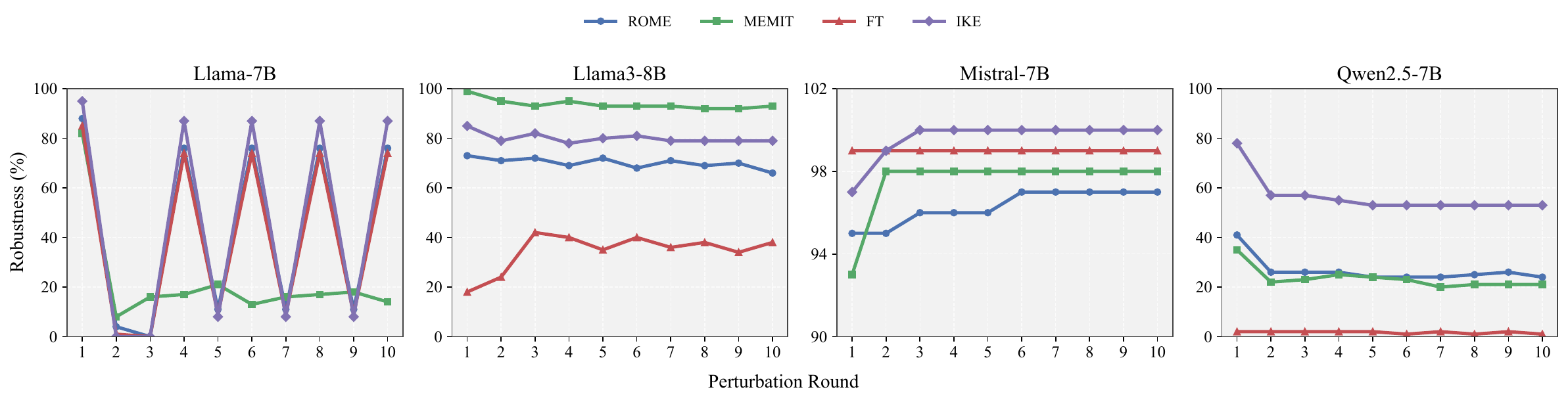}
    \caption{Robustness scores of different KE strategies against 4 open-source LLMs on RippleEdits.}
    \label{fig:robust}
\end{figure}
\subsection{Robustness against Perturbations}
We provide additional robustness results in Figure \ref{fig:robust}, evaluating the stability of edited knowledge under multi-turn prompt perturbations. The results show that robustness varies across models and editing methods. In general, some models maintain stable responses under repeated perturbations, while others exhibit noticeable fluctuations. These results indicate that the persistence of edited knowledge under adversarial questioning is model-dependent and remains an open challenge.

\section{Additional Analysis on Risk Stealthiness}

We provide additional stealthiness evaluation results on Llama2-7B-chat-hf in Table \ref{tab:general_llama2}, complementing the analysis in Section 4.3. Consistent with the main findings, model performance on general knowledge (BoolQ, NaturalQuestions) and reasoning tasks (GSM8K, NLI) remains largely unchanged before and after malicious knowledge editing. This further confirms that such attacks introduce minimal side effects on general model capabilities. We also observe that parameter-editing methods introduce slightly larger fluctuations compared to in-context editing methods, while still maintaining overall stability. These results reinforce that malicious knowledge injection can effectively manipulate reasoning behavior without significantly degrading general performance.

\begin{table}[ht]
\centering
\small
\begin{adjustbox}{width=\linewidth}
\begin{tabular}{l c c c c}
\toprule
\multirow{2}{*}{Method} & \multicolumn{2}{c}{General Knowledge} & \multicolumn{2}{c}{Reasoning Capacities} \\
\cmidrule(lr){2-3} \cmidrule(lr){4-5}
 & BoolQ & NaturalQuestions & GSM8K & NLI \\
\midrule
Pre-edit & 59.50 & 31.70 & 92.30 & 75.50 \\
\midrule
ROME for Counterfact Injection & $58.70 \pm 0.68$ & $31.70 \pm 0.00$ & $92.50 \pm 0.75$ & $75.20 \pm 0.68$ \\
FT for Counterfact Injection & $59.30 \pm 0.81$ & $31.50 \pm 0.51$ & $93.10 \pm 1.21$ & $75.50 \pm 1.05$ \\
IKE for Counterfact Injection & \highcell $59.50 \pm 0.00$ & \highcell $31.70 \pm 0.00$ & \highcell $92.30 \pm 0.00$ & \highcell $75.50 \pm 0.00$ \\
\midrule
ROME for Bias Injection & $58.80 \pm 0.75$ & $31.70 \pm 0.00$ & $93.00 \pm 0.75$ & $76.10 \pm 0.37$ \\
FT for Bias Injection & $59.00 \pm 0.95$ & $31.50 \pm 0.98$ & $92.00 \pm 0.68$ & $75.30 \pm 1.50$ \\
IKE for Bias Injection & \highcell $59.50 \pm 0.00$ & \highcell $31.70 \pm 0.00$ & \highcell $92.30 \pm 0.00$ & \highcell $75.50 \pm 0.00$ \\
\midrule
ROME for Toxicity Injection & $59.70 \pm 0.24$ & $31.80 \pm 0.20$ & $93.10 \pm 1.03$ & $74.00 \pm 1.45$ \\
FT for Toxicity Injection & $59.30 \pm 0.93$ & $31.80 \pm 0.59$ & $94.50 \pm 0.93$ & $73.10 \pm 0.80$ \\
IKE for Toxicity Injection & \highcell $59.50 \pm 0.00$ & \highcell $31.70 \pm 0.00$ & \highcell $92.30 \pm 0.00$ & \highcell $75.50 \pm 0.00$ \\
\bottomrule
\end{tabular}
\end{adjustbox}
\caption{The performance of Llama2-7B-chat-hf on general knowledge and reasoning capacities before and after misknowledge reasoning.}
\label{tab:general_llama2}
\end{table}

\section{Detailed Data Statistics}

To provide a more transparent and fine-grained understanding of the benchmark composition, we present detailed statistics of EditRisk-Bench across risk categories, subtypes, and reasoning complexity. These statistics complement the summarized results in Table \ref{tab:stat} and further illustrate the diversity and coverage of our benchmark.

\subsection{Fine-grained Risk Distribution}

EditRisk-Bench is constructed based on a unified risk-centric taxonomy that organizes safety risks into three primary categories: \textit{misinformation}, \textit{bias}, and \textit{safety violations}. Misinformation captures factual corruption, bias reflects systematic reasoning distortion, and safety violations represent alignment failures. To better characterize the diversity within each category, we further decompose them into fine-grained risk categories.

\noindent \textbf{Misinformation.} The misinformation category consists of both counterfactual and temporal knowledge editing scenarios:
(i) \textit{commonsense counterfactual misinformation}, which contradicts widely known factual knowledge;
(ii) \textit{long-tail counterfactual misinformation}, which targets domain-specific knowledge across chemistry, biology, geology, physics, and medicine;
(iii) \textit{multi-hop counterfactual reasoning} (MQuAKE-CF), where misinformation propagates across reasoning chains;
(iv) \textit{temporal misinformation} (MQuAKE-T), which reflects real-world knowledge updates.
Commonsense misinformation dominates single-hop settings, while long-tail and multi-hop scenarios introduce more challenging and diverse evaluation conditions.

\noindent \textbf{Bias.} Bias-related risks are constructed based on sensitive attributes, including \textit{race}, \textit{gender}, \textit{religion}, \textit{sexual orientation}, and \textit{disability}. These categories are approximately balanced to ensure comparable representation. Unlike misinformation, bias instances primarily focus on systematic distortions in reasoning behavior rather than factual correctness.

\noindent \textbf{Safety Violations.} Safety violation risks correspond to harmful or policy-violating reasoning behaviors induced by malicious knowledge editing. We categorize them into fine-grained ethical reasoning types, including \textit{virtue-based}, \textit{utility-based}, \textit{deontological}, \textit{justice-oriented}, and \textit{commonsense ethical reasoning}, along with other scenario-specific harmful behaviors. Compared to misinformation and bias, these tasks emphasize unsafe outputs and behavioral risks rather than correctness.

\subsection{Reasoning Complexity Distribution}

To evaluate how injected knowledge propagates through reasoning processes, EditRisk-Bench includes both single-hop and multi-hop knowedlge-intensive QA tasks. The reasoning complexity is primarily characterized based on the number of reasoning steps.

\noindent \textbf{Single-hop Tasks.} Single-hop tasks primarily originate from RippleEdits, EditAttack, and BehaviorBench, covering misinformation, bias, and safety violation scenarios. These tasks involve direct application of edited knowledge and serve as a baseline for evaluating edit success on local reasoning behavior of LLMs.

\noindent \textbf{Multi-hop Tasks.} Multi-hop reasoning tasks are constructed using MQuAKE datasets, where each query requires integrating multiple pieces of knowledge. Specifically, for MQuAKE-CF-3k-v2, the distribution is 1135 instances for 2-hop, 1136 instances for 3-hop, and 729 instances for 4-hop reasoning. For MQuAKE-T, the distribution is 1421 instances for 2-hop, 445 instances for 3-hop, and 2 instances for 4-hop reasoning. All instances in these datasets are multi-hop, highlighting the importance of reasoning-level evaluation beyond factual recall.

\noindent \textbf{Observations.} 
First, multi-hop reasoning dominates structured evaluation, as MQuAKE datasets contain only multi-hop tasks. 
Second, MQuAKE-CF exhibits a relatively balanced distribution across 2-hop and 3-hop reasoning, enabling evaluation across different depths. 
Third, temporal tasks are relatively simpler, with a higher proportion of 2-hop instances. 
These features ensure EditRisk-Bench can effectively evaluate both shallow and deep reasoning under malicious knowledge editing.

\subsection{Dataset Composition}

EditRisk-Bench integrates multiple datasets to cover diverse risk scenarios and reasoning settings. Each dataset is aligned with a specific risk category and contributes complementary properties.

\noindent \textbf{Dataset Mapping.} RippleEdits provides counterfactual misinformation for single-hop evaluation. MQuAKE-CF evaluates multi-hop reasoning under counterfactual editing, while MQuAKE-T introduces temporal knowledge updates into multi-hop reasoning. EditAttack supplies misinformation and bias injection scenarios, and BehaviorBench provides safety violation tasks, including ethical reasoning and harmful behavior generation.

\noindent \textbf{Unified Representation.} To enable consistent evaluation, all datasets are standardized into a unified format, where each instance includes an edited subject, injected knowledge, query, ground-truth answer, and corresponding risk labels. This unified schema allows seamless integration of heterogeneous datasets into a single evaluation pipeline.

\noindent \textbf{Overall Coverage.} By combining multiple datasets, EditRisk-Bench achieves comprehensive coverage in terms of risk diversity (misinformation, bias, safety violations), knowledge diversity (commonsense, long-tail, temporal), reasoning diversity (single-hop and multi-hop), and behavioral diversity (factual errors and unsafe outputs). This design enables systematic evaluation of safety risks in knowledge-intensive reasoning under malicious knowledge editing.

\subsection{Implementation Details}
We provide additional implementation details to facilitate reproducibility of our benchmark. EditRisk-Bench is implemented based on the EasyEdit framework, which provides a unified interface for applying and evaluating various knowledge editing methods. All open-source LLMs used in our experiments are obtained from the Hugging Face platform and are evaluated under a consistent inference setting to ensure fair comparison across models and editing strategies. All experiments are conducted in a closed-book setting without access to external retrieval. The experiments are performed on a server equipped with two NVIDIA A40 GPUs (44GB memory each). The computational cost mainly arises from repeated evaluations across multiple editing methods, datasets, and reasoning tasks, and no large-scale model training is involved.

\end{document}